\documentclass[11pt]{diazessay} % Font size (can be 10pt, 11pt or 12pt)

\usepackage{authblk}
\usepackage{graphicx} % Required for inserting images
\usepackage{csquotes}
\usepackage{soul}
\usepackage{float}
\usepackage{hyperref}

\title{Rejecting Cognitivism: Computational Phenomenology for Deep Learning}

\author[1]{Pierre Beckmann}
\author[1]{Guillaume Köstner}
\author[2,3]{Inês Hipólito}
\affil[1]{Université de Lausanne (UNIL), Department of Philosophy}
\affil[2]{Humboldt-Universität zu Berlin, Department of Philosophy \& Berlin School of Mind and Brain}
\affil[3]{University of Amsterdam, Faculty of Social and Behavioural Sciences }
\date{}                     %% if you don't need date to appear
\begin{document}

\maketitle % Print the title section

%----------------------------------------------------------------------------------------
%	ABSTRACT AND KEYWORDS
%----------------------------------------------------------------------------------------

%\renewcommand{\abstractname}{Summary} % Uncomment to change the name of the abstract to something else

\begin{abstract}
We propose a non-representationalist framework for deep learning relying on a novel method: computational phenomenology, a dialogue between the first-person perspective (relying on phenomenology) and the mechanisms of computational models. We thereby reject the modern cognitivist interpretation of deep learning, according to which artificial neural networks encode representations of external entities. This interpretation mainly relies on neuro-representationalism, a position that combines a strong ontological commitment towards scientific theoretical entities and the idea that the brain operates on symbolic representations of these entities. We proceed as follows: after offering a review of cognitivism and neuro-representationalism in the field of deep learning, we first elaborate a phenomenological critique of these positions; we then sketch out computational phenomenology and distinguish it from existing alternatives; finally we apply this new method to deep learning models trained on specific tasks, in order to formulate a conceptual framework of deep-learning, that allows one to think of artificial neural networks’ mechanisms in terms of lived experience.
\end{abstract}

\vspace{30pt} % Vertical whitespace between the abstract and first section

%----------------------------------------------------------------------------------------
%	ESSAY BODY
%----------------------------------------------------------------------------------------

\newenvironment{hangingpar}[1]
  {\begin{list}
          {}
          {\setlength{\itemindent}{-#1}%%'
           \setlength{\leftmargin}{#1}%%'
           \setlength{\itemsep}{0pt}%%'
           \setlength{\parsep}{\parskip}%%'
           \setlength{\topsep}{\parskip}%%'
           }
    \setlength{\parindent}{-#1}%%
    \item[]
  }
  {\end{list}}

\section{Introduction}

In the past years, deep learning (DL) has achieved impressive feats, its
artificial neural networks (ANNs) competing with human performance on
tasks involving the understanding or the generation of text (Devlin et
al., 2019; Brown et al., 2020; Schulman et al., 2022), images (Radford
et al. 2021; Ramesh et al. 2021) or speech (Baevski et al. 2020; Hsu et
al., 2021). DL has had a revolutionary impact in the industry and
society, as well as in scientific research (DeVries et al., 2018; Davies
et al., 2021; Jumper et al., 2021). Specifically, ANNs ability to
successfully mimic a number of human cognitive processes, has fueled
comparative research between DL and cognitive sciences or neurosciences
(Yamins, 2016; Kumar et al., 2022; McClelland, 2022; Millet et al.,
2022). Recently, DL also motivated renewed philosophical perspectives
upon old questions relating to the mind, brain and behavior (Buckner,
2019; Sloman, 2019; Fazi 2021; Perconti \& Plebe, 2020).

Despite its successes, ANNs are notoriously hard to interpret, in the
sense that we cannot exactly understand how they solve their tasks
(Boge, 2022). For this reason they are sometimes referred to as ``black
boxes'' (Castelvecchi, 2016). This opacity makes DL models susceptible
to diverse interpretations through different conceptual frameworks. The
most prominent framework for the interpretation of DL has been
cognitivism, the first research program in cognitive sciences (MacKay et
al., 1956; Lees \& Chomsky, 1957; Minsky, 1961). Relying on the
functioning of the Turing machine, cognitivism defends cognition in
terms of \emph{symbol} manipulation: cognitive processes are thought to
rely on \emph{representation} of entities\footnote{We use the
  ontologically neutral term of ``entity'' as it does not matter for our
  purpose whether the ontology of the world is conceived in terms of
  objects, properties, relations, processes, events, tropes, or other
  metaphysical categories. What is important is the representative
  relation holding between those entities and the mental symbols
  according to cognitivism. It should also be noted that this relation
  does not imply that the grammar of those symbols has to match the
  ontology of the world. For example, the representation \emph{of} a
  process does not need to represent it \emph{as a process}.} of an
external pregiven world. This approach has been influential both in
cognitive sciences and philosophy of mind, with Fodor's computational
theory of mind being the most prominent (Fodor, 1983).

DL originates within connectionism, a computationalist framework that
disputes that cognitive, computational processes are leveraged by symbol
manipulation (McCulloch \& Pitts, 1943; Rosenblatt, 1958; Rumelhart et
al., 1986). Aiming for a more ``biological'' resemblance of the
distributed operations of the brain, connectionism brings forth the
ancestors of DL models, such as Rosenblatt's perceptron (Rosenblatt,
1958). Unlike cognitivism's Turing machine, these mathematical models do
not need to be implemented fully: they can learn how to solve tasks by
slowly adjusting to new inputs. With these models, connectionism
promotes a new conceptual framework to think about human cognition as
\emph{emergence} of global \emph{states} to fulfill cognitive functions.
It therefore initially opposes cognitivist's symbol-like representations
: ``one-to-one mappings'' between entity and representation (Rosenblatt,
1958)\footnote{However, as Bechtel (1991) notes, from the 1980s,
  connectionists try to reconcile their theory with the predominant
  cognitivism, insisting that their models ``should be embraced as a
  subsymbolic alternative to symbolic models of cognition''.}.

Although it stems from connectionism, DL is today largely thought in
terms of \emph{representation}-based operations -- relying on the
cognitivist toolkit. In fact, connectionist models' self-organizational
capabilities happen to be of use for the convinced cognitivist, because
they can provide the otherwise unanswered explanation of how a system
can \emph{learn} symbols. ANNs learn symbolic representations of
external properties on the basis of which they can execute further
computations to solve tasks. This stance is quickly associated with
neuro-representationalism (NR), combining a strong realism towards
scientific entities with the idea that we experience a brain generated
model of these entities (Churchland and Sejnowski, 1990; Milkowski,
2013; Mrowca et al., 2018; Sitzmann et al., 2020).

NR with the notion of representation is pervasive both in computational
neuroscience (Piantadosi, 2021; Poldrack, 2021) and DL itself (LeCun et
al., 2015; Ha \& Schmidhuber, 2018; Gidaris et al., 2018; Chen et al.,
2020; Goh et al., 2021; Matsuo et al., 2022). In \emph{Nature}'s most
cited paper, ``Deep Learning'' for example, ANNs are described from the
first paragraph, as machines that can ``be fed with raw data'' and
``automatically discover the \emph{representations} needed for detection
or classification'' (LeCun et al., 2015). To the best of our knowledge,
this framework is implicitly prevalent in deep learning literature.
Often, deep learning researchers motivate their use of the cognitivist
concept of representation by relying on NR; see this recent deep
learning textbook for example:

\begin{quote}
More often than not, hidden layers have fewer neurons than the input
layer to force the network to learn compressed representations of the
original input. For example, while our eyes obtain raw pixel values from
our surroundings, our brain thinks in terms of edges and contours. This
is because the hidden layers of biological neurons in our brain force us
to come up with better representations for everything we perceive.
(Buduma et al., 2022)
\end{quote}

Furthermore, DL researchers commonly interpret ANNs as learning ``world
models'', that mimic external world structures and dynamics to plan
ahead (Ha \& Schmidhuber, 2018; Matsuo et al., 2022). When used to
understand the mind, these ``world models'' are oftentimes reduced to
perception; the idea being the following: because perception is
indirect, the brain must build internal models in an attempt to
represent what could potentially be the perceptual space state of
affairs (Von der Malsburg, 1995; Ashby, 2014; Saddler, Gonzalez \&
McDermott, 2022). This introduction of intermediate representations
posits the existence of an external reality, the ontological structure
of which can (arguably) be known independently of the way the mind
relates to it (metaphysical realism), but that we can never directly
access through our perception (representationalism). In philosophy, this
form of representationalism is famously opposed by phenomenology, which
puts on hold the question of the existence of an external reality in
favor of a rigorous description of lived experience (Husserl, 1931;
Merleau-Ponty, 1945). Today, in cognitive science, and under the
influence of phenomenology, representationalism is further challenged in
the Embodied and Enactive Cognitive Science (EECS) research program
(Varela et al, 1991; Hutto \& Myin, 2012; Chemero, 2011; Di Paolo et
al., 2017; Gallagher, 2017).

Taking into consideration these critiques of representationalism, this
paper aims to provide a conceptual framework of DL that does not rely on
symbols as the basic units of cognition. As such, we chose to rely on
phenomenology, privileging insights from the careful description of
reality as it appears to us in first-person perspective. Furthermore,
following Lutz and Thompson (2003), we will leverage three levels of
enquiry, or sources of exploration of cognition : \emph{the
neurophysiological source}, \emph{the phenomenological source} and
\emph{the computational source}. Historically, the computational source
was used to formalize and link the findings obtained from the two other
sources. With DL successfully mimicking some of our cognitive processes,
the computational source now generates new \emph{data} and becomes a new
source of exploration. This observation opens up the possibility of
\emph{computational phenomenology (CP)}: an exclusive dialogue between
the \emph{computational} and the \emph{phenomenological} source that
puts on hold the question of the material basis of cognitive processes
(which belongs to the neurophysiological source). The point of such a
dialogue is not to disqualify the neurophysiological source, but rather
to provisionally let the two other sources free of any constraint or
import coming from the third one. As such, this dialogue is more
faithful to phenomenology (relying on first-person descriptions of
experience) than EECS approaches that tend to recast phenomenology in a
more naturalist, third-person point of view. Turning to DL, we find that
from lived experience, the apparent non-decomposability of ANN
operations -- their ``black box'' aspect -- is not surprising as the
underlying mechanics of many of our cognitive processes are unclear, or
opaque, to us. This observation allows us to both propose new
phenomenology-drawn concepts to think of ANN operations, and to embrace
the opaque processual nature of cognitive processes. The
contribution of this paper is threefold, we propose computational
phenomenology, a new methodology and conceptual framework for
philosophers and cognitive scientists to conceive (1) of the mind and
its relation to (2) task-solving computational models; such a conceptual
framework provides (3) DL engineers with a new experience-based toolkit
by applying this methodology to the operations of ANNs.

\section{Cognitivism and neuro-representationalism in deep learning}

DL is concerned with the design and the training of ANNs. An ANN
sequentially connects \emph{layers} of (non-linear) threshold-activated
nodes with linear operations according to a set of \emph{weights} to
transform an input into an output. Layers between the input layer and
the output layer are called \emph{hidden layers}; the ``deep'' in DL
refers to the multiple hidden layers in the ANNs.

The weights of an ANN are not fixed but are gradually adjusted to better
solve a precise task. To enable learning, the deep learning researcher
picks three main ingredients, given some particular data: the network's
\emph{architecture} (the way in which the different nodes are connected
through the weights), a \emph{loss function} (that models the objective
of the task) and a \emph{learning rule} (that dictates the way the
weights are updated according to the loss function -- and is almost
always based on an algorithm called \emph{backpropagation} (Rumelhart et
al., 1986). Once the optimization of the weights -- also called
\emph{training} -- is done, ANNs allow \emph{inference}: the propagation
of new input all the way to output layers, causing the nodes of the
hidden layers to \emph{activate} in a particular way. In short, once the
ANN is trained, the \emph{weights} are fixed and allow the processing of
new inputs; on the other hand, the \emph{activations} are obtained
through inference and always correspond to one given input. These
weight-determined successive activations, or patterns of activations,
are not easily interpretable. Importantly, this means that it is not
possible --~so far at least -- to clearly decompose them into
explainable steps, or symbol-based operations. The original
connectionist framework sees them as nothing more than \emph{emerging
states} and is already satisfied that they allow solving a particular
cognitive task (Varela et al., 1991, p. 98). However, some researchers
reject the apparent weaker explanatory power of this interpretation in
favor of reading of DL in terms of cognitivist's mind-computer metaphor
: i.e. conceiving cognition as symbolic computation (Shinozaki, 2021;
Zbontar et al., 2021; Matsuo et al., 2022; Ceccon et al., 2022; Leek,
2022).

A cognitivist reading of an ANN relies on interpreting its pattern of
activations as symbol-based operations. We argue that this reading isn't
motivated by technical reasons but that it is grounded in a
philosophical worldview. Cognitivism, from its birth in the 1950s,
relies on the mechanisms of the Turing machine to interpret human
cognition (Putnam, 1967; Fodor, 1983). When one implements a Turing
machine, one decides what a one or a zero in a given cell \emph{means,}
that is to say, to what it relates to in our world. This physically
manipulable entry then carries a human-assigned meaning; it becomes a
\emph{symbol} that \emph{represents} a given external entity, and on the
basis of which Turing machines can carry out meaningful operations. In a
second step, cognitivism transposes this functioning to the mind. The
mind is thought to run on the basis of \emph{mental}
\emph{representations}, that semantically \emph{encode} properties of an
external pregiven world analogously to the symbols of a Turing machine.
Cognitivism therefore implies a \emph{metaphysical realism}, the view
that there exists an external world, with entities, independently of our
perception or thoughts of it. Cognitivism can be systematized around the
answers given to three major questions:

\begin{enumerate}
\item
\ul{What is the world?}

An external reality that exists independently of our cognition of it.

\item
  \ul{What is a representation?}

A symbol that stands for an entity of the world.

\item
  \ul{What is the mind?}
  
A system capable of rule-based operations to carry out cognitive
processes.
\end{enumerate}

Metaphysical realism is however a philosophical assumption or
standpoint, which cannot be proven by cognitivism: after all, by
cognitivist lights, we only ever have access to our mental
representations of the external world. Furthermore, in this setting it
isn't clear how the correspondence between the entity and the
representation is grounded. There is no possible way to step outside our
mental representations and observe the real world and the way it relates
to our representations of it in the same way the implementer of a Turing
machine knows what symbol relates to what entity.

As a mathematical model capable -- after training -- to solve tasks
without relying on the assignment of symbols (except for input and
output nodes), the ANN is of special interest to cognitivism. Rather
than questioning her conceptual framework -- given that computation
without representation suddenly seems possible --, the convinced
cognitivist supposes that ANNs do in fact rely on internal symbols to
solve tasks. Henceforth, she obtains a model that does not need symbol
assignment as it \emph{learns} them. Following this position, ANNs offer
cognitivism's missing element and can be thought of as some sort of
elaborate Turing machine that self-adapts to obtain symbol-like
representations of some external properties. They are thought to
simultaneously learn how to \emph{represent} important properties --
\emph{for} themselves (as if they were the implementer of the Turing
machine) -- and how to use these representations to solve their task.
This stance however relies on the supposition that ANNs do indeed learn
Turing-like symbols\footnote{At this point the reader might argue that
  the fact that ANNs can be implemented on Turing machines does show
  that they do rely on symbols. The reason this argument fails is that,
  in such a case, the numbers on the tape of the Turing machine don't
  explicitly stand for anything, since it was not the implementer who
  chose what they stand for.}. Furthermore, it is not clear if
cognitivism's initial problem is solved as even by relying on the ANN it
is not clear how our representations relate to the external world.

In today's cognitive sciences and deep learning literature, this idea of
a self-organizing Turing machine is frequently taken up in a new version
of cognitivism: neuro-representationalism (NR). In NR, cognitivism's
metaphysical realism takes the specific form of a strong scientific
realism: the external entities that exist independently of us become the
theoretical entities of modern natural sciences. Additionally the
cognitivist-interpreted ANN becomes a model of the brain; in other
words, the brain is thought to learn Turing-like mental representations
of the world. Finally, NR takes cognitivism's representationalism to its
extreme by making a claim about our conscious experience: we experience
a brain generated model of these entities (Churchland and Sejnowski,
1990; Mrowca et al., 2018; Sitzmann et al., 2020); for criticism see
(Zahavi, 2018; Hipólito 2022). Frith puts it in slogan form: ``my
perception is not of the world, but of my brain's model of the world''
(Frith, 2007). According to NR, agents do not directly perceive a photon
and its wavelength but only a mental representation of it -- in this
case, a certain color (Metzinger, 2009). NR too can be systematized
around three major questions:

\begin{enumerate}
\item
\ul{What is the world?}

The physical world, composed of the scientific entities discovered by
science.
\item
  \ul{What is a representation?}

A symbol learned by the brain that encodes an external entity of the

\item
  \ul{What is the mind?}
  
Composed of the cognitive processes responsible for mental phenomena --
which consist in neural computations on the basis of the representations
-- and of the content of these phenomena, consciousness -- which is
given only by the representations.
\end{enumerate}

NR's concept of representation is pervasive in the DL field. In
\emph{Nature}'s most cited paper, ``Deep Learning'', we find a common
cognitivist interpretation of ANNs from the first paragraph, describing
them as machines that can ``be fed with raw data'' and ``automatically
discover the \emph{representations} needed for detection or
classification'' (LeCun et al., 2015). Following this view, the first
layers of the ANN operate a \emph{feature extraction}, computing
relevant features, or representations, while the last layers adequately
combine these -- carrying out some form of ``reasoning''. It is also
common to consider the layers as extracting features hierarchically,
where the first layers compute ``low-level'' representations -- such as
edges in an image -- and subsequent layers compute ``high-level''
representations -- such as larger motifs in an image (LeCun et al.,
2015). An analogous interpretation can be found in computational
neurosciences, when describing, for example, early vision as extracting
low-level representations on the basis of which higher-level reasoning
is carried out in the brain (LeCun et al., 2015). As mentioned in the
introduction this implicit framework is a given in modern deep learning
research and frequently specifically relies on NR and its understanding
of human cognition.

\begin{figure}[h]
\centering
\includegraphics[width=0.8\textwidth]{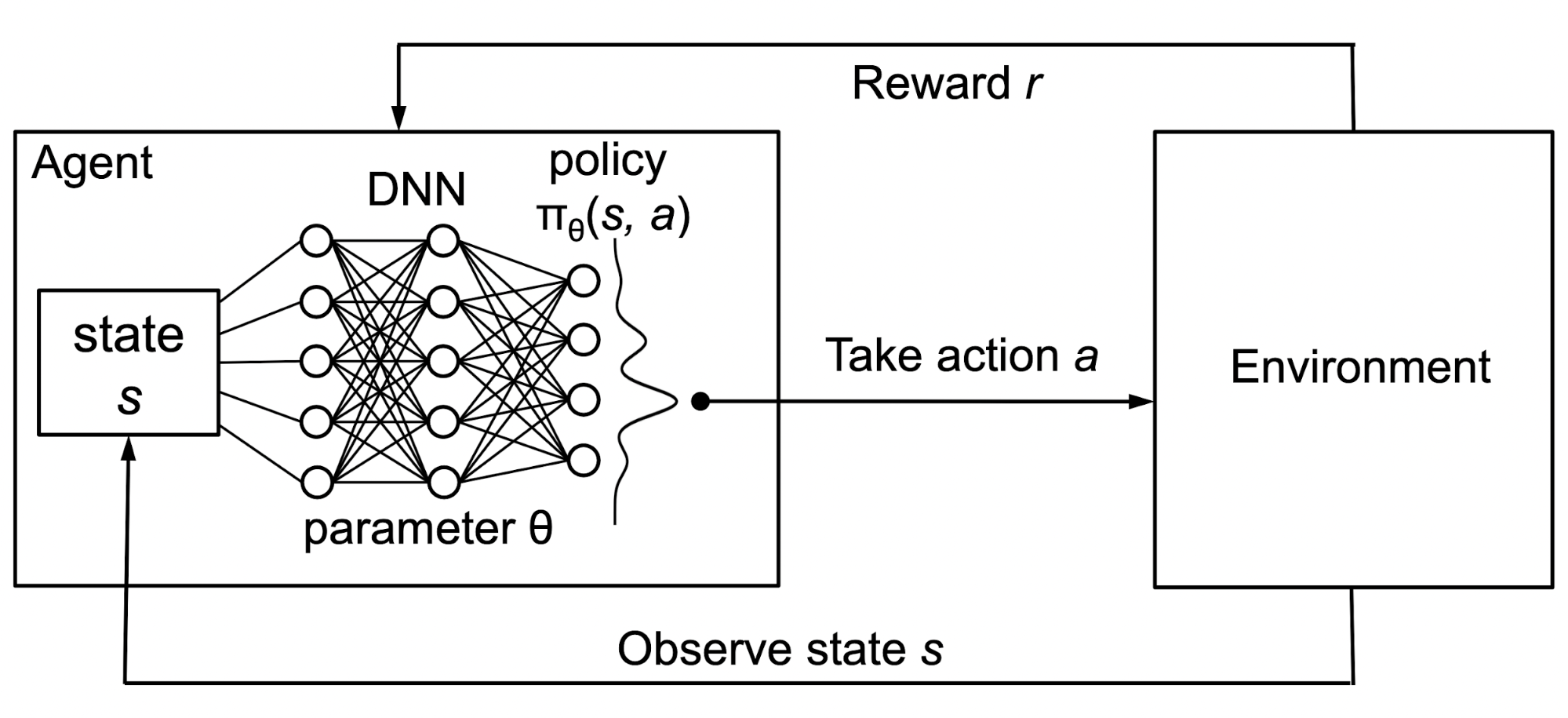}
\caption{Model-free deep reinforcement learning. The ANN, or
DNN (deep neural network), learns to pick an action based on
observations of an external environment, in order to maximize reward.
Taken from (Mao et al., 2016)}
\label{fig:RL1}
\end{figure}

\begin{figure}[h]
\centering
\includegraphics[width=0.55\textwidth]{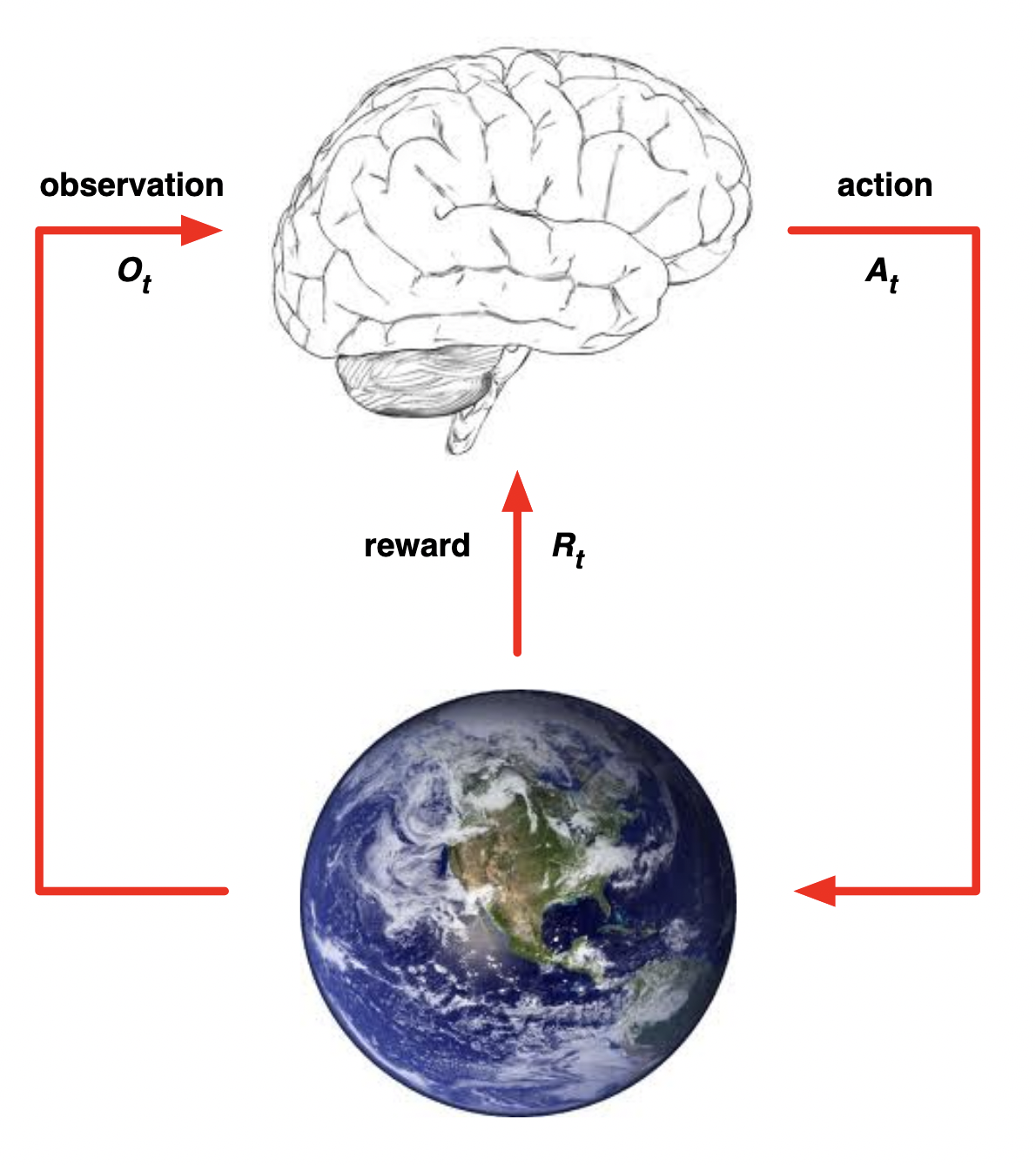}
\caption{Figure from deep mind's introduction to reinforcement
learning course (Silver, 2015). To introduce deep reinforcement
learning, the brain-in-world analogy is used to explain the situation of
the ANN in its environment.}
\label{fig:RL2}
\end{figure}

A fundamental condition for the acceptance of the construct of NR is the
existence of a pre-given world. Reenacting such an agent/world setting
corresponds to a large field of DL called deep reinforcement learning
(DRL) (Mnih et al., 2015; Silver et al., 2016; Eppe et al. 2022; Wang et
al, 2022). DRL trains as ANN, considered as an agent, to select the
right actions based on the observations (or states) of an external
environment in order to maximize potential reward (Fig. \ref{fig:RL1}). In this setting the ANN is typically thought to learn an internal \emph{world model} that \emph{represents} the dynamics of the environment and allows it to predict, reason, and plan (Ha \& Schmidhuber, 2018; Goyal et al.,
2022; Mazzaglia et al., 2022; Driess et al. 2022). The concept of
``world model'' is very important among researchers: it is often thought
to be the key element to human-level intelligence (Matsuo et al., 2022).
While, in the literature the term ``world model'' can refer to different
things, in the context of DL it means that the ANN does not rely
directly on the external environment but operates on the basis of its
``world model''. Once again, the motivation comes from the cognitivist
assumption that an agent does not have direct access to the world and
therefore has to ``wonder'' and ``interact'' with it through the
mediation of internal representations. If this internal character is
taken as far as making the world model the realm of perceived things, we
get the key ingredient for NR: conscious experience, then, is (limited
to) our model of the world. Subsequently, when the ANN learns to extract
features from its environment, it is thought to be analogous to the
brain learning to ``come up'' with internal representations of the
hidden physical world. By this reasoning, DRL becomes a direct
mathematical formalization of NR. In the Machine Learning research
community, DRL -- a growing field in neuroscience (Botvinick et al.,
2020) -- is almost exclusively interpreted in this NR setting -- in
fact, the brain-in-the-world analogy is typically used to introduce the
field of DRL (Fig. \ref{fig:RL2}).

In short, although DL stems from connectionism, a program compelled to
reject the challenges that come with assuming (Turing) symbol
computation mechanisms on the neurobiological level, the majority of
research today carried out in DL still incorporates the cognitivist
toolkit, employing concepts (such as world models, representation, etc.)
which, then, returns cognition as reducing to NR.

The DL researcher might however object that the concept of
representation is of technical use in numerous applications. Indeed, the
term representation is often used as a synonym of the term
\emph{embedding}, a low-dimensional feature typically obtained by
extracting node activations at a particular layer of a trained ANN. When
working with images for example, an engineer can extract the
pen-ultimate layer's activations of an ANN trained on a first task and
use it as a representation, or embedding, on a downstream task such as
image classification (Chen et al., 2020). Because this procedure works,
the DL engineer might argue that ANNs do indeed rely on representations.
This line of thought can be challenged. First of all, these embeddings
correspond to one given datum (one precise image for example), not a
certain general property of the data. Secondly, they are representations
that the engineer chooses for downstream tasks, they are representations
for the scientist; they do not stand one-to-one for the datum -- in
fact, no perfect inverse mapping, that would allow retrieving the datum
from an embedding alone, is technically possible. The technical utility
of embeddings only shows that patterns of activations at a particular
level contain relevant material -- that might be, in essence,
non-decomposable -- to solve a number of tasks. Should this reasoning
hold, embeddings -- just as the activations of an ANN -- are hardly
representations in the sense of Turing machine symbols.

In conclusion, little technical or empirical reasons plead in favor of
the employment of Turing-like representations \emph{for} \emph{the ANN}.
While in some cases it can be established that some layers are more
sensitive to particular patterns (for example edges in an image), the
weights and activations of ANNs are notoriously hard to interpret (Zhang
et al., 2021). Even when some interpretation is possible (Olah, 2015;
Goh et al., 2021), the particular patterns could simply represent
particular properties of some data \emph{for} \emph{the
engineer/scientist} that interprets them given a certain task (Boge,
2022). In any case, if some form of interpretation is possible
decomposition into symbol-based operations seems nowhere near. This is
why we will henceforth employ the concept of ANN ``non-decomposability''
-- as in non-decomposable into symbol-based operations -- rather than
vaguer ``uninterpretability'' that is sometimes used in DL literature.

In the next sections we will propose a conceptual framework for DL that
doesn't rely on the cognitivist notion of representation. We will do so
by first taking some insights from phenomenology and then applying them
to deep learning.

\section{The phenomenological critique of cognitivism: implications for Deep Learning}

As seen in the previous section, NR takes a robust scientific realist
stance, i.e. existence of an external pre-given objective world (that
can be described by science and represented by cognitive processes);
from then on, the question of consciousness becomes: how does it arise
from natural processes? This is notoriously difficult and referred to as
the hard problem of consciousness (Chalmers, 1995).

Phenomenology flips the problem around\footnote{It is impossible for us
  to offer an all-encompassing account of the tremendously rich
  phenomenological tradition or to engage with exegetical issues
  pertaining to the thought of Husserl and other important thinkers in
  said tradition. We will limit ourselves to highlighting the points
  that are essential for the approach we want to sketch in these pages.
  For detail, see (Zahavi, 2008; Gallagher and Zahavi, 2020).}. Instead
of positing the existence of an outside world and questioning the
emergence of consciousness, phenomenology seeks to describe how the
world appears to us in lived phenomena (Merleau-Ponty, 1945). It puts on
hold any question regarding the existence of the external world
(\emph{i.e.}, whether something is \emph{really} out there or if we are
simply hallucinating our reality), its ontology (\emph{i.e.}, the types
of things that really exist), and the mind-world relationship (idealism,
realism, \emph{etc}.). Husserl (1931, §32: 59-60) calls this bracketing
of judgment ``\emph{epoché}'', which aims at neutralizing what he calls
the ``natural attitude''. This attitude is characterized by a
common-sense belief in the reality of external, discrete, ordinary
objects. Merleau-Ponty (2012: p. 69) calls this natural attitude
``objective thought'' and interprets it as the shared assumption of
idealism and realism, and as the unquestioned metaphysics of modern
natural sciences\footnote{It is important to note that phenomenology is
  not \emph{per se} incompatible with a moderate scientific realism
  which claims that natural sciences discovers real objective features
  of the world and produces true statements about it (see, for example,
  Dreyfus 1992), although it rejects the idea that the scientific image
  of the world is a complete and exhaustive descriptive-explanatory
  framework.}. Once every judgment pertaining to the natural
attitude/objective thought has been put into brackets, phenomenology
takes as a starting point a passive stance with regards to phenomena.
That is, it lets things appear as they appear spontaneously to the mind
that is directed towards the world without trying to categorize them.
The task of phenomenology is \emph{then} to describe the structures of
manifestation, producing an understanding of the mind and its
interactions with the world that differs strongly from the views exposed
in the previous section. One of the most fundamental features of the
mind that phenomenology emphasizes is \emph{intentionality}, that is the
fact that mental states are \emph{directed towards} something. For
example, an episode of perception is always a perception \emph{of}
something. Following Husserl (1900; 1931, §37), we will call the thing
towards which the mind is directed the ``intentional object'', and the
conscious mental state directed towards the intentional object an
``intentional act''.

Phenomenology and NR diverge significantly in the way they describe
cognitive processes. For NR, which operates on a clear separation
between subject and external world, a cognitive process can be cast as a
sequence of symbol-based operations, from perception (sense data inputs)
to a particular action (motor output), or storage of a new useful
representation. With its bracketing, phenomenology considers cognitive
processes from a different point of view where it makes no sense to
distinguish an external entity from our representation of it; there are
simply intentional objects that appear to me: consciousness and the
world are given in one stroke. Therefore, cognitive processes are not
considered as an algorithmic processing of perceptual inputs, but rather
as habits that underlie and structure our lived experience, ``that
simultaneously {[}delimit{]} our field of vision and our field of
action'' (Merleau-Ponty, 1945, p. 153).

Merleau-Ponty (1945, p. 143) offers a useful illustration of the problem
posed to those conceiving of perception as indirect: the blind man's
cane. He depicts and opposes the position of the intellectualist (which
seems identical to NR's position), according to which the blind man
infers the shape of external objects in two steps: first, by deducing
the cane's position given its pressure on the hand and then, by
inferring the shape given this position (Merleau-Ponty, 1945, p. 153).
He argues that once that the blindman gets used to the cane, once he has
it ``in hand'', the habit precisely ``relieves {[}him{]} of this very
task'' (Merleau-Ponty, 1945, p. 153). The cane becomes ``an instrument
\emph{with} which he perceives'', its tip is ``transformed into a
sensitive zone'', expanding his perceived world (Merleau-Ponty, 1945, p.
154). The acquired cane-sensing skill is simultaneously a perceptual
habit and a motor habit (there is no perception without movement of the
cane) that structures conscious experience. It grounds an ``organic
relation between the subject and the world'' that does not rely on
symbol-like representations (Merleau-Ponty, 1945, p. 154).

Could this different perspective on cognition open up a different way to
interpret computational models? Dreyfus, drawing upon Heidegger's
phenomenology, considers cognitive processes as acquired habits (and
consequently opposes cognitivism's representationalism). He argues that
we do not acquire skills by storing representations but by a gradual
refinement of our perception that offers new solicitations in given
situations in the world; therefore, ``the best model of the world is the
world itself'' (Dreyfus, 2007). He, for example, rejects the existence
of an internal map: ``what we have learned from our experience of
finding our way around in a city is `sedimented' in how that city
\emph{looks} to us'' (Dreyfus, 2007, p. 1144). Dreyfus insists that
basic cognitive processes (he gives the examples of driving a car or
playing chess) do not rely explicitly on symbols but are just the result
of a gradual adaptation -- they are representation-less (Dreyfus, 2002).
Therefore, he sees the advent of ANNs as a strong blow against
cognitivism's commitment to representations as they ``provide a model of
how the past can affect present perception and action without the brain
needing to store specific memories at all'' (Dreyfus, 2002, p. 374).
From the phenomenological standpoint, ANNs seeming non-decomposabilty is
particularly interesting because it evokes the opacity of our implicit
habits (i.e. that do not seem to rely on representations).

However, the use of the term ``brain'' in Dreyfus's previous citation
marks an important conceptual shift that shouldn't go unnoticed.
Important observations that were acquired from the phenomenological
source are mapped onto the functioning of the brain -- a system
understood in terms of and by scientific investigation that belongs to
the neurophysiological source. Dreyfus's implicit supposition of an
overlap between the phenomenological and the neurophysiological is in
tension with phenomenology's initial ambition to put on hold the
question of existence of external objects. And this supposition can be
taken a step further by trying to reduce the phenomenological to the
natural -- shifting from cognition explained from first-person
perspective to cognition explained from third-person perspective.

This \emph{naturalization} of phenomenology is constitutive of Varela's
formulation of enactivism, proposed as an alternative to cognitivism
(Varela et al., 1991). His enactivism retains all major principles
uncovered by the phenomenological source and uses them to think the
coupling processes of an agent with its environment, considered in a
naturalistic setting. It deems that, through sensorimotor activity,
organisms become structurally coupled to their environment which allows
them to \emph{enact} a world. The ``organic relation'' between
consciousness and world from first-person perspective becomes a
``structural coupling'' between organism and environment (Varela et al.,
1991, p. 206). When considering the relation between mind and
computational models, this enactivism is interested, for example, in
robots that can navigate autonomously in the external world; Varela et
al. (1991) for example turn towards Brooks's finite state machines
(Brooks, 1987). This brings it closer to the understanding of cognitive
processes as a way to interact with the external world, rather than
understanding them as habits that underlie our lived experience. Because
enactivism aims to bridge phenomenology and neurophysiology, it operates
upon a third-person separation between agent and environment;
phenomenology doesn't rely on this dissociation and considers
experiences (that are structured by habits) in which the I and the world
are intertwined. We therefore insist on the necessity of a clear
epistemic separation between the phenomenological source and the
neurophysiological source (in the tradition of Husserl's epoche) to
approach the question of the relationship between cognition and
computation. This distinction opens up the possibility of establishing
parallels between cognitive processes seen as \emph{habits} and
computational models in a way that is more faithful to phenomenology's
initial project. Parallels that will be useful to develop a
representation-less toolkit for DL.

\section{Computational phenomenology and deep learning}

In the previous section, we have highlighted the need for a method that
considers computational models from phenomenology without reducing it to
structures uncovered by third-person sciences, such as by
neurophysiology\footnote{Note that mathematical formalizations tend to
  be seen only (possibly wrongly) as tools to describe objective natural
  processes. Phenomenology has therefore traditionally been skeptical of
  any attempts of mathematical formalization. Husserl for example,
  famously described Galileo as ``at once a discovering and a concealing
  genius'' because his mathematization of natural phenomena ends up
  covering up its origin: lived phenomena (Husserl, 1936, p. 53).}. The
first such method that distinguished phenomenology, neurophysiology and
mathematical/computational modeling is neurophenomenology (Varela,
1996). Neurophenomenology aims to establish ``reciprocal constraints''
between first-person data from phenomenology and measured data of
physiological processes, in order to allow a dialogue, or a
``circulation'', between the internal (phenomenological) and external
(scientific) accounts of a given cognitive process (Varela, 1996, p.
343); a third component is then used to provide a ``neutral ground''
between ``these two kinds of accounts: formal (or computational) models
(Lutz \& Thompson, 2003). Some approaches specifically focus on
mathematically formalizing the structures of lived experience directly,
referred to as the ``CREA proposal'' (Petitot \& Smith, 1996; Petitot
1999). However their obtained mathematical formalizations correspond to
a ``physical behavior'' from which the ``qualitative structure of a
phenomenon'' \emph{emerges} (Petitot \& Smith, 1996, p. 241); the
computational model describes neurophysiological processes from which
phenomenological events emerge (similar to connectionism).

More recently, Yoshimi (2011) proposed to associate phenomenological
structures with neuro-computational structures, considering
connectionist models. The first approach to propose computational models
of the structures of lived experience \emph{without} necessarily
assuming that they belong to the neurophysiological is Ramstead et al.'s
(2022) computational phenomenology. This active inference-inspired
approach seeks to build generative models that explain data from ``our
first-person phenomenology \emph{itself}'' (Ramstead et al., 2022). In a
second step, such models (of perception, or of action for example) can
also be used in simulations (Sandved-Smith et al., 2021).

What all these approaches have not accounted for fully is the
``trainability'' aspect: computational models can now be \emph{trained},
and therefore generate their own kind of data, in the same way as the
neurophysiological and the phenomenological can. In our computational
phenomenology account, we propose to use the term \emph{source} to
designate a distinct level of inquiry that can provide its own type of
data\footnote{A source is therefore defined epistemologically and not
  ontologically: it doesn't designate a level of reality but a set of
  methods from which we can obtain a specific type of data. We owe the
  term to Bitbol (2006).}. Following the three fields of knowledge of
Lutz \& Thompson (2003), we therefore propose to distinguish three
different \emph{sources} to explore cognition: \emph{the
neurophysiological source}, \emph{the phenomenological source} and
\emph{the computational source} -- insisting that computational models
are not simply tools to formalize data from other fields of knowledge
but that they can generate their own.

In fact, neuroscience researchers have long ago recognized the
computational as a source of exploration for cognition. Notably, the
advent of this new source has strengthened the tendency of
neuroscientific research to cast aside the phenomenological source by
implicitly assuming a close correspondence between physiological
processes and conscious experience. In DL-based computational
neuroscience, numerous research completely dismisses the
phenomenological source and compares ANN processes to brain processes
(comparison which does seem quite natural given DL's initial structural
inspiration being the brain); examples include Yamins' paradigmatic
research on vision in brains and in ANNs (Yamins \& DiCarlo, 2016),
other recent investigations (Kumar et al., 2022; Millet et al., 2022)
and even proposed research programs (Doerig et al., 2022; Cohen et al.,
2022).

\begin{figure}[h]
\centering
\includegraphics[width=0.8\textwidth]{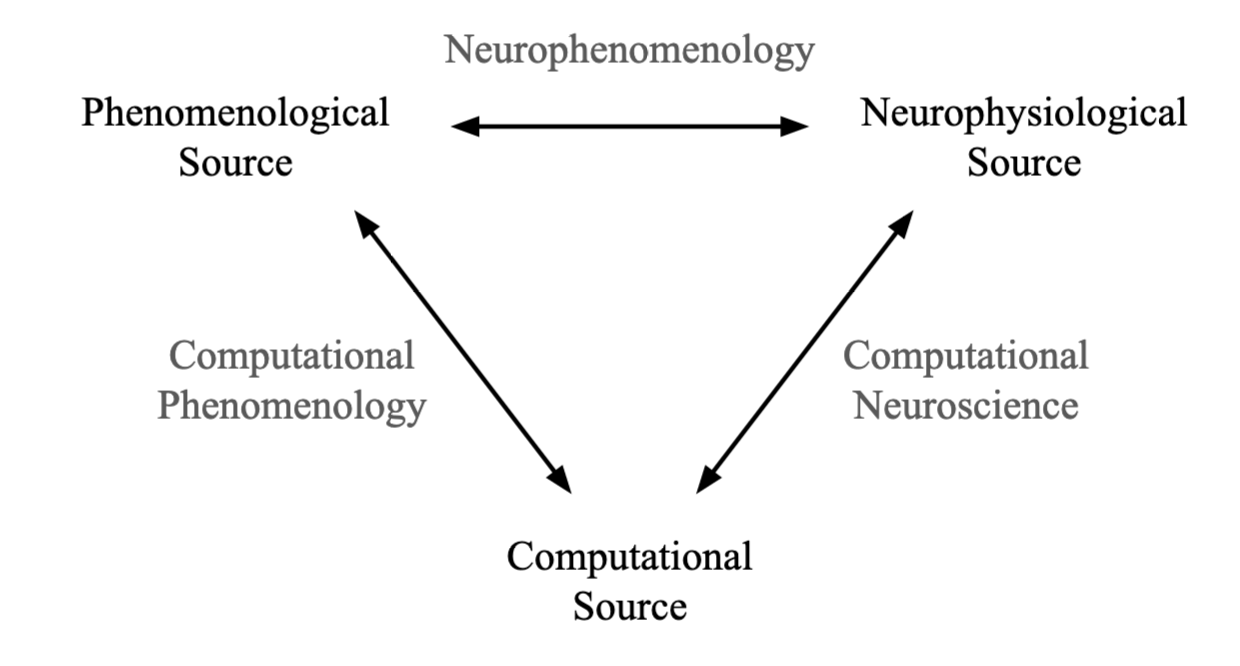}
\caption{The three sources of exploration of cognition. Modern
computational sciences promote a circulation of knowledge between
neuroscience and computational sciences ; in consequence they tend to
exclude the phenomenological aspect of cognition. In response, we
propose computational phenomenology, an epistemic dialogue between the
phenomenological and the computational, in a similar vein than
neurophenomenology, a dialogue between the neurophysiological and the
phenomenological.}
\label{fig:three_sources}
\end{figure}

To restore the phenomenological source and give it an epistemic role in
DL research, we propose a new formulation of \emph{computational
phenomenology} (CP), defining it as an epistemic dialogue between the
phenomenological source and computational source (in an analogous way to
neurophenomenology). Specifically, we consider circulation of knowledge
between phenomenological descriptions of a given cognitive process (and
its corresponding habit) and the mechanisms of trained computational
models on a corresponding task. In line with the phenomenological
tradition, we bracket the question of the existence of the external
world and provisionally let the two sources free of any constraint or
import coming from the neurophysiological source. Therefore CP doesn't
help itself to the third-person separation between agent and environment
(which neurophenomenology as well as enactivism rely on) and considers
cognitive processes, and corresponding habits, in which consciousness
and the world are intertwined. CP's goal is to uncover structural
invariants and guiding constraints between phenomenology and
computational modeling. However, it also provides a ground to develop a
new framework to think computational models different from the existing
ones (cognitivism, connectionism or NR)\footnote{Our CP approach can be
  considered a ``cousin'' of Ramstead et al.'s active inference
  formulation (2022). The main difference being that Ramstead' et al.s
  CP mathematizes the underlying structures of experience directly,
  whereas our version identifies common mechanisms between AI systems
  and corresponding first-person experiences of cognitive processes in
  the perspective to formulate an alternative to cognitivism.} (See Fig. \ref{fig:three_sources}.).

Once the computational is recast as a \emph{source} (because it
generates data), it makes sense to select the models that yield the most
impressive results, that are best at solving cognitive tasks. Therefore
it is natural for CP to turn towards DL. In the next sections, we will
engage in a dialogue between phenomenology and deep learning
systematically covering important dimensions of cognition: perception,
action, imagination and language. We will see that ANNs
non-decomposaibilty is particularly interesting when considering
phenomenological descriptions. However, we will not limit our
investigation to highlight a shared opacity in DL and in phenomenology.
We will rely on common DL interpretation techniques\footnote{For a
  review of such interpretation techniques see (Räuker et al., 2022).}
to highlight similarities between specific cognitive processes, cast as
habits, and ANNs trained on an analogous task. This investigation will
notably allow us to redefine the term of \emph{representation}.
Specifically, we abandon the idea of symbol-like representations
underlying our cognition and will characterize the decomposable
phenomenal content of our cognitive processes -- that means, our
conscious representations. In what follows, we offer clear directions
within our account of computational phenomenology to its employment within specific DL/cognition areas: perception (4.1), imagination (4.2),
and language (4.3).

\subsection{Merleau-pontian perception: a new setting in which to consider learning}

Merleau-Ponty's phenomenology gives a central role to the body
(Merleau-Ponty, 1945). In his theoretical framework, the functional unit
that allows the imbrication of consciousness and world through action
and perception is the \emph{corps propre} (one's own body). The
\emph{corps propre} is not the body considered as an object, but the
body I live as, and through which I have a world. It is the body I
discover by investigating the underlying structures of the first-person
perspective; as opposed to the biological body discovered by ``objective
thought'' that is ``unaware of the subject of perception''
(Merleau-Ponty, 1945, p. 214). It does not rely on the concept of brain,
as the brain cannot be identified as the origin of perception from the
lived phenomena themselves just as ``it would be absurd to say that I
see with my eyes or that I hear with my ears'' (Merleau-Ponty, 1945, p.
214). When it comes to explaining cognition, the biological body
(including the brain) is the functional unit of the neurophysiological
source whereas the \emph{corps propre} is the functional unit of the
phenomenological source. Considering perception by modeling the former
as an ANN tends to fall into NR. An alternative is given by
investigating the similarities between the latter and DL.

It turns out the ANN is a good candidate in modeling the way the
\emph{corps propre} adjusts its grip on the world. The \emph{corps
propre} enables learning from past experiences by \emph{sedimentation}
into \emph{habits}. Merleau-Ponty uses this term to stress that
experiences aren't stored separately, as symbols in a Turing machine or
entries on a hard disk, but rather consolidate into new habits in an
analogous way solid material settles (sediments) at the bottom of a
liquid. The seemingly uninterpretable way the particles settle resembles
the adjustment of weights during the training of an ANN. The sedimented
habits form a ``contracted knowledge'' that isn't ``an inert mass at the
foundation of our consciousness'' but is ``taken up'' in every ``new
movement'' towards the world (Merleau-Ponty, 1945, p. 132) in the same
way that at inference, all the weights of an ANN are mobilized to
propagate the content of each input. In accordance with ANNs that do not
store all the examples in the form of representations but rather benefit
from them by weight adjustment, the \emph{corps propre}'s intentional
acts are not based on a superposition, but a sedimentation of
experiences. Intentional acts allowed by the \emph{corps propre} include
the \emph{perceptual synthesis}: the detachment of a privileged object
from an indifferent background. As such our perceptions are grounded on
a perceptual tradition, that contracts the history of previous
experiences in a way that remains wholly opaque to the subject of
perception -- opacity, or non-decomposability, that is also
characteristic of ANNs.

That perception is grounded on the sedimentation of past experiences
means that what we learn is directly transcribed in the way the world
shows up to us. The adjustment of perception corresponds to the search
for an \emph{optimal grip} on the lived world that allows confident
action. This means that the features of the world that appear to us do
not correspond to some encoded properties of external realities. This
finding turns some of NR's most elementary assumptions completely on
their head: the space that I experience in perception, for example, is
now conceived as the result of my grip on the world. The orientation of
the whole of our perception -- what we consider as being ``top'',
``bottom'', ``left'' and ``right'' -- at any given time, translates a
certain equilibrium I reached in my lived world, rather than the
encoding of some universally given directions.

But how does this redefinition of perception -- as a dynamically
adjusted foundation of our experiences rather than a faithful
reconstruction of external entities -- allow a dialogue with DL models?
Let's consider two related cases where perception is perturbed by
rotation of our visual field. When I look at an upside-down face for
some time, it becomes ``monstrous'': I see a ``pointed and hairless
head'' with a ``blood-red orifice'' on its forehead and ``two moving
eyeballs'' where the mouth should be (Merleau-Ponty, 1945, p. 263). My
lack of interaction with inverted faces translates into an unstable
\emph{grip}. An ANN trained to recognize faces that would be presented
to it systematically the "right side up", would fail if they were
suddenly turned around. In such cases, the neural network is said to
have difficulty generalizing, i.e., adapting to a new type of data.
Analogously to the \emph{corps propre}, an ANN is constantly trying to
adjust its ``grip'' on the data, and doesn't rely on a translation into
symbols -- that, in this case, would have allowed it to instantly revert
the inflicted rotation.

\begin{figure}[h]
\centering
\includegraphics[width=0.9\textwidth]{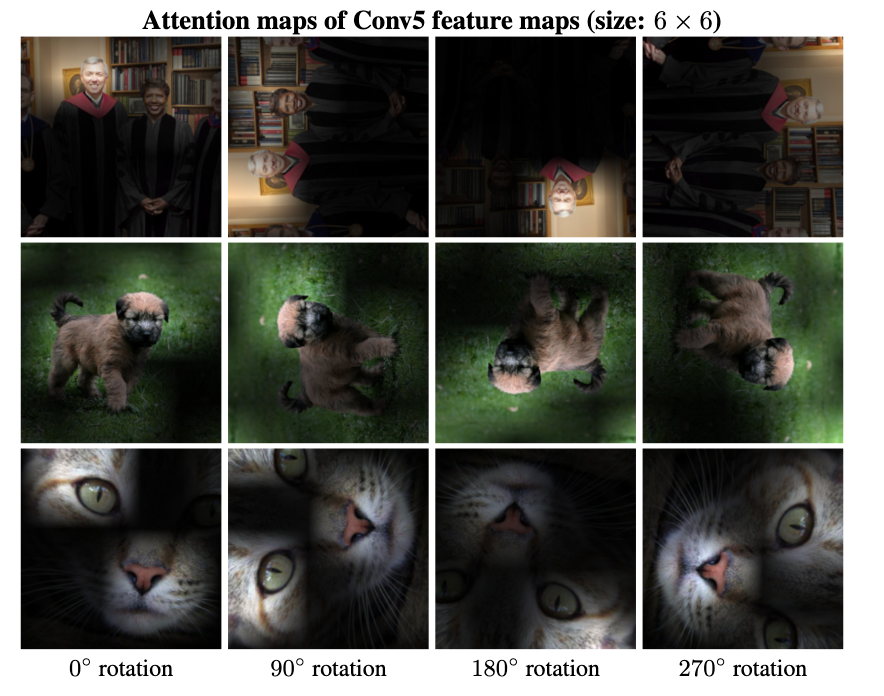}
\caption{Attention maps of the artificial neural network Rotnet that learns to recognize the angle of rotation of an image (Gidaris, 2018).}
\label{fig:rotnet}
\end{figure}

\begin{figure}[h]
\centering
\includegraphics[width=\textwidth]{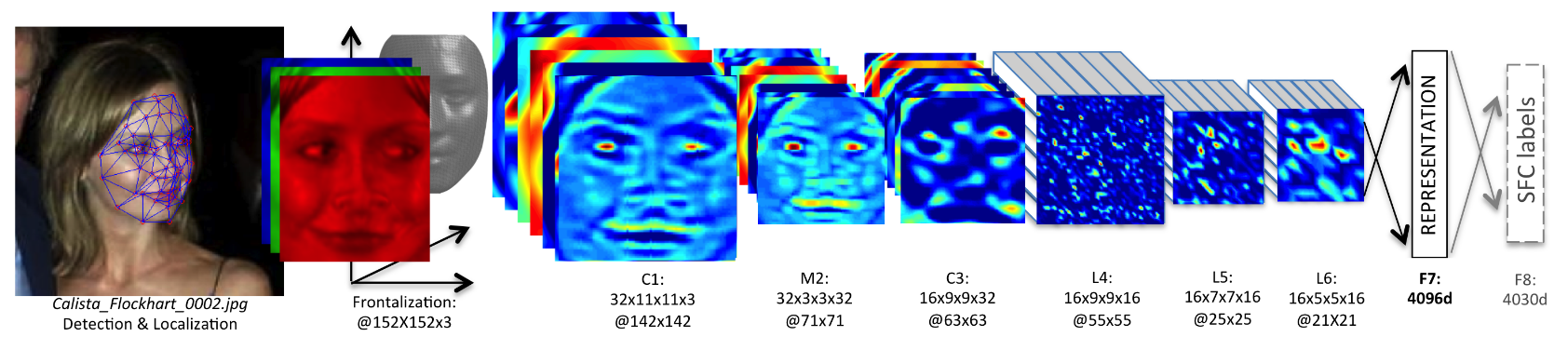}
\caption{Architecture of the \emph{DeepFace} model and activations
corresponding to a given input. First blue square from the left is the most telling (Taigman,
2014).}
\label{fig:deepface}
\end{figure}

The \emph{corps propre} can also readapt in cases of spatial level
shifting. In one of Wertheimer's experiments, a subject observing his
room for a few minutes through a mirror that tilts it by 45 degrees,
suddenly sees his visual spatial level shift when he projects himself
into this new setting. The ``spectacle'' offered by the tilted room is a
call to a new ``virtual'' \emph{corps propre}. That is, the body with
``the legs and arms that it would take to walk and act in'' the room, to
open the cupboard or sit at the table (Merleau-Ponty, 1945, p. 289). At
that point perception adjusts in a way to guide actions in this new
setting: the spatial level shifts. What exactly triggers this tilting?
Before the shift, the orientation of the (intentional) objects in the
mirror is not natural and does not allow the usual interaction with
them. It is therefore the objects -- and particularly faces, according
to Merleau-Ponty -- that are the sign of a tilted visual field and serve
as anchor points to switch to a new oriented phenomenal space in which
they are the ``right side up''. It turns out an ANN learns a similar
``trick'' to identify rotated images: \emph{RotNet}, trained to identify
the rotation angle (0, 90, 180, or 270 degrees) of rotated images, also
relies on the orientation of objects and faces present in the scene
(Gidaris et al., 2018). Indeed, attention maps -- which overlay the
input image with network activations -- reveal a high attention
attributed to faces in rotation angle classification (Fig. \ref{fig:rotnet}). The orientation of the objects themselves, which allows them to serve as
anchor points, is given by a ``perceptual itinerary'', a learned order
in which I visit its important features. \emph{DeepFace} (Taigman et
al., 2014), a neural network that learns to identify a specific person
from their face, also relies on certain learned key points as shown by
higher activations at pixels corresponding to the eyes and mouth (Fig.
\ref{fig:deepface})\footnote{However, these key points are not visited in a certain
  order. The ANNs inference lacks the temporal dimension of our lived
  perceptual synthesis.}.

We have first highlighted similarities between the \emph{corps propre}'s
grip adjustment and the training of an ANN; we have then shown a case
where an ANN and corps propre learn a similar ``trick''. But what do
these correspondences actually show? First, it is interesting to note
that simple mathematical functions fitted with a simple learning rule
(in this case, gradient descent) end up relying on similar tricks than
the \emph{corps propre} to solve analogous tasks. Furthermore, they show
how computational learning can be conceptualized in the lived world, in
conscious experience, rather than with respect to an external world.
Indeed, the input to the spatial level adjustment process are not some
photons or other scientific entities but our previous lived experience
(which calls for an adjustment if it does not allow an optimal grip).
Thinking these kinds of processes in terms of ANNs paves the way for a
new framework to conceive DL, in which learning happens in the lived
world. Furthermore, ANNs' non-decomposability is good news in this
setting because it opens up the idea that some processes could, by
nature, be non-decomposable into operations on symbols -- in the same
way adjustment processes are \emph{opaque} to the subject of perception.
They allow us to interpret cognition from the phenomenological source
without relying on unconscious representations. As for conscious
representations (phenomenal contents), we have seen with the case of the
orientation of space or of intentional objects, that they are not the
encoding of some external entity but the result of an optimal grip on
the world. We will seek to further characterize them in the two
following sections.

\subsection{Sartrean imagination: conscious re-presentation of sedimented experiences}

If ANNs are thought to model processes of the \emph{corps propre}, their
training is recast as a \emph{sedimentation} of past experiences in
order to obtain an \emph{optimal grip} on the \emph{lived world} by
refining the \emph{perceptual synthesis} which determines the way things
appear to us. But how can the world be ``the best model of itself'' when
I \emph{imagine} something? Do we not store some representations of
things to be able to later form mental images of them? Indeed,
cognitivism might turn to this form of explanation relying on the
Turing-machine's functioning, positing that we save some representations
to be able to later access them. This explanation corresponds to Hume's
understanding of mental images as faint copies of past perceptions. In
\emph{The imaginary}, Sartre rejects Hume's copy principle because it
falls under what he calls the ``illusion of immanence''. He holds that
mental images are not copies \emph{contained} in consciousness and
recasts them as \emph{imaging consciousnesses} (Sartre, 2004). If Hume's
copy principle is best explained with the Turing-machine, we will show
that Sartre's \emph{imaging consciousness} better resembles the cascade
of activations of an ANN.

Sartre rejects the idea of the mental image as a copy; in fact, he
considers it as incompatible with the dynamic nature of consciousness:
indeed, it would be ``impossible to slip these material portraits into a
conscious synthetic structure without destroying the structure, cutting
the contacts, stopping the current, breaking the continuity.'' (Sartre,
2004, p. 6). The act of imagining something cannot rely on a symbol, or
any object it would be heterogeneous to; rather, the act of imagining
and the mental image are one. ``The majority of psychologists
{[}mistakenly{]} think that they find the image in taking a
cross-section through the current of consciousness'', Sartre says
(Sartre, 2004, p. 15); this error is repeated by anyone trying to
interpret DL using the cognitivism framework by trying to find
representation in the ``current'' of successful activations of an ANN
inference. Therefore, we should rethink the mental image -- that is an
\emph{imaging consciousness} -- as the complete inference of an ANN.

What does it mean for a mental image to be an imaging consciousness? And
why cannot it be a representation that would be retrieved by
consciousness? Just like the \emph{perceptual synthesis}, the
\emph{imaging consciousness} is an intentional act. And in both
processes, the object is aimed at as a corporeal object while never
\emph{entering} my consciousness. Indeed, when I perceive the chair, it
would be absurd to say that the chair \emph{enters} into my
consciousness; in the same way an \emph{imagining consciousness} does
not rely on a copy of a chair that would be \emph{in} my consciousness.
Both intentional acts do not rely on separable symbols. However, even if
it relies on a perceptual tradition, the \emph{perceptual synthesis}
``encounters'' the object it aims at (Sartre, 2004, p. 7). This does not
seem to be the case for the \emph{imagining consciousness}. On what
knowledge can it rely? How can it aim at the sensitive elements of
objects it first encountered in perception?

\begin{figure}[h]
\centering
\includegraphics[width=0.95\textwidth]{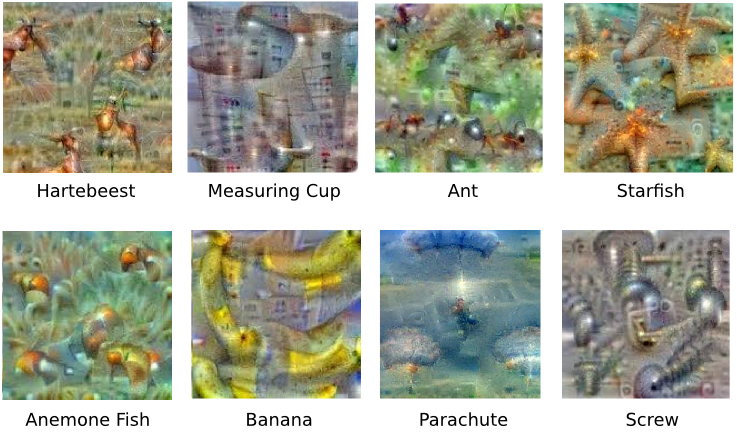}
\caption{\emph{DeepDream} images that maximize activations
corresponding to particular image labels. (Mordvintsev, 2015).}
\label{fig:deepdream}
\end{figure}

To picture something as possessing certain sensible qualities, to aim at
my friend as ``blond, tall, with a snub or aquiline nose, etc.'', my
\emph{imaging consciousness} ``must aim through a certain layer of
consciousness that we can call the layer of knowledge'' (Sartre, 2004,
p. 57). From Sartre's descriptions, we can formulate a hypothesis
according to which the \emph{imaging consciousness} re-employs certain
tracks of the \emph{perceptual synthesis}. In DL mechanisms, this could
mean that the imaging consciousness uses the sedimented weights of
processes belonging to the \emph{perceptual synthesis} (such as the
process that allows the orientation of an object). In fact, there exists
a technique in DL -- called \emph{DeepDream} -- that allows one to
generate images from a trained image classification network: for
example, by tweaking random noise in order to maximize the activation of
a final output node corresponding to one particular label (Mordvintsev
et al., 2015) (Fig. \ref{fig:deepdream}). Recently, (Fei et al., 2022) have employed a
similar technique to visualize what they call the ``imagination'' of a
model trained to bring closer an image and a matching textual
description in a higher-level feature space. In a similar way, my
\emph{imaging consciousness} could try to strongly re-activate the
processes that allow me to recognize (in perception) my friend as being
``blond, tall, with a snub or aquiline nose, etc.''. This, of course,
would be only one possible way to do it; the general idea we want to
convey is that past perceptual experiences might \emph{sediment} into
perceptual \emph{habits}, and that in a second phase some structures, or
some ``regions'', of these habits might be used, or \emph{re-activated},
by \emph{imaging consciousnesses} (which are themselves are particular
type of habits). As such, an imagined object isn't really red because we
stored its color in symbol form, it is red because we employ a certain
red-making process (that relies on the process used to recognize objects
as red in perception): we imagine \emph{redly}.

Once the mental image recast as an \emph{imaging consciousness}, Sartre
establishes two characteristics of the aimed object in the way it
appears to us. These also apply, analogously, to images generated by DL.
Mental images are given as an ``undifferentiated whole'', meaning that
they can possess qualitative properties but be quantitatively undefined:
it is possible for example to have a mental image of the Pantheon but to
be unable to count the columns -- it can have an undetermined number of
them (Sartre, 2004, p. 87). Furthermore, they do not obey the
``principle of individuation'': when I imagine my friend for example,
she can appear simultaneously from the front and the profile. Both the
qualitative undefinition and the superposition of various viewpoints can
be observed in images generated by \emph{DeepDream} (Fig. \ref{fig:deepdream}) or other DL
image generations techniques (Fei et al., 2022; Ramesh et al., 2021).
These properties, or failure cases, of imagination from the
phenomenological source can be observed analogously in DL based image
generation. The TM, on the other hand, cannot explain an
``undifferentiated'' Pantheon; it is more in line with Hume's copy
principle, where one either has a stored copy of the Pantheon and can
retrieve or one does not and cannot output anything. Dedicated DL models
better explain borderline cases of mental images considered from the
phenomenological source than the TM, cognitivism's go-to model, does.

In conclusion, Sartre's concept of \emph{imaging consciousness} can be
modeled by the inference of an ANN taken as a whole process, wherein one
does not look for any symbol-like representations. It relies on
previously \emph{sedimented} perceptions allowing it to
\emph{re-present} (show again) properties aimed at by the
\emph{perceptual synthesis} by \emph{re-activating} them. Examples of
such properties include the color and even the orientation of an object.
We use the term \emph{re-presentation} to stress that previous
experiential content is \emph{re-activated}, rather than symbolically
represented: we imagine \emph{redly} instead of accessing the symbol red
in our internal database. Therefore, our analysis of imagination
culminates in a new understanding of phenomenal content as
\emph{conscious re-presentations}.

\subsection{Language: concepts as conscious re-presentations}

Computational phenomenology also has to address a major aspect of
cognition that is also an important research domain in deep learning:
language. Language and concept-based reasoning heavily rely on
individuated units or conscious symbols. These units provide the ability
to group and recognize things by subsuming them under the same concept.
These concepts can be employed even in the absence of any instance of
it, for example in a propositional act. In such a case a concept is
linked with a series of sounds, to a word. These ``meaningful cores''
can also be explored from the phenomenological source: with
Merleau-Ponty, a word is a ``phonetic gesture'', generated through my
body and its vocal cords, that ``produces a certain structuring of
experience'' in the same way than my perceptual habits do
(Merleau-Ponty, 1945, p. 199). As they crystallize into words, concepts
become communicable: in the same movement ``the body opens itself to a
new behavior and renders that behavior intelligible to external
observers'' (Merleau-Ponty, 1945, p. 199). We can consider concepts as
\emph{conscious re-presentations} by extending the sartrean conception
of imagination.

As we did for perception and imagination, we could consider concepts as
being cognitive processes that rely on conceptual \emph{habits}. This
thesis can be mapped to ANN mechanisms : the inference of the ANN can
stand for the concept as a cognitive process (or \emph{conscious
re-presentation}), the training -- or sedimentation of past experiences
-- can stand for the formation of the \emph{conceptual} \emph{habit}.
Similarly to mental images concepts might also reemploy some sedimented
tracks of other habits (perceptual or imaginative\footnote{Sartre for
  example speaks of ``illustrations'' and ``symbolic schemas'', that are
  two different kinds of mental images that can accompany our
  recollection of a particular concept (Sartre, 2004, p. 87).}). Here
particular ANNs can be used to reason about different functions of
concepts understood as cognitive processes. Panaccio proposes two
fundamental roles of the concept: its \emph{representational role} and
its \emph{inferential role} (Panaccio, 2011).

The \emph{representational role} refers to a concept's relationship to
perception allowing recognition and re-presentation. The concept
\emph{red} allows me to perceive objects as red in my lived world and
also to re-present \emph{redly} in imagination. As such concepts, and
associated words, are structuring lived experience; they are not name
tags of external properties. Merleau-Ponty stresses this point by
analyzing a study of Gelb and Goldstein where patients with color name
amnesia were incapable of grouping objects of the same color
(Merleau-Ponty, 1945, p. 197). From DL, models of image classification
-- where a concept (a word) is associated with an image -- would then be
of interest to investigate the \emph{representational role} of concepts.
In such cases, the task of image recognition sediments into weights
which determine the pattern of successive activations at inference
(perception) and allow re-activation (imagination). The aspect of
experience-structuring can also be observed analogously: a recent DL
study, for example, showed that semantic segmentation emerges when
training a network to caption images (Xu et al., 2022).

The \emph{inferential role} refers to a concept's relationship to
language. Inherent to my concept of a \emph{cat} is the ability it gives
me to draw inferences from it: ``a cat is a mammal'', ``a cat has four
legs'', etc. Interestingly, I can possess only the inferential dimension
of a concept: even if I never saw a platypus, and even if I have no idea
what it looks like, I might still know that it is a mammal. In such
cases, we first encounter the word and then form the corresponding
concept based on the things we learn it implies (Panaccio, 2011, p. 61).
To model the \emph{inferential role} of concepts drawing from deep
learning, we can turn to DL language modeling (Brown et al., 2020;
Devlin et al., 2019). Under masked language modeling, sentences are fed
word by word to an ANN; however, certain words are randomly masked, and
the task of the model is to predict them. Once trained, the ANN allows
the generation of highly coherent text. Recent DL research has shown
promising results for approaches relying on both of these concept roles
(Fei et al., 2022; Li et al., 2020; Radford et al., 2021) -- two roles
that most of the concepts of everyday use play simultaneously.

This last section on mental language and language allows us to
generalize our proposal. In the lived world we can isolate some
cognitive processes that structure our experience, for example that give
the objects their orientation or allow me to recognize them as falling
under the concept of \emph{red}. These processes continuously rely on a
\emph{sedimentation} of past experiences, which allows an \emph{optimal
action-orienting grip} on the world. \emph{Habits}, resulting from the
sedimentation, also allow re-presentation, in the form of imagination or
recollection: I can form mental images and I can draw inferences from my
concepts (even those I only ``encountered'' through language and not
through direct perception). These re-presentations can be thought of as
re-activations of sedimented weights of an ANN. Finally, in all these
processes, conscious representation (or concept) and cognitive process
are one: cognitive processes are not to be decomposed into symbols and
their content is to be thought of as a process and not as an object.
Both the mechanism of an ANN (processual) and its seeming
non-decomposability support this thesis.

\section{A new toolkit for deep learning and a novel mathematization of cognition}

Computational phenomenology is a novel approach to work towards a
computational interpretation of cognitive processes as described from
first-person perspective. Rather than mathematizing cognition considered
as emerging from the external world as described by science, it takes a
new background of exploration: conscious experience. The CP recipe is
the following: (1) place oneself in the lived world, (2) isolate a
particular cognitive process and (3) compare it to the mechanism that a
computational model uses to solve an analogous task. We believe CP has
potential for practical applications in both phenomenology and deep
learning: it could provide the phenomenologists some clues as to what
aspects to investigate given the way an ANN solves a given task (like
the mechanism of RotNet indicates that faces might help to determine any
tilt of the perceptive field); and it could provide the deep learning
engineer some new ideas to design DL setups (for example by trying to
generate images using the weights of trained recognition networks in an
analogous way to the mechanism of sartrean imagination). From
phenomenology to deep learning, one could even hope for some insights to
tackle some more fundamental aspects of modern DL -- could be
investigated: a first-person-perspective credible learning rule (an
alternative to \emph{backpropagation}) or a way to make the inference of
an ANN more temporal (like the habits of the \emph{corps propre}
are)\footnote{This aspect was raised in footnote 10: the perceptual
  synthesis (one of the habits of the \emph{corps propre}) is temporal
  -- we constantly adjust what we see in time -- whereas the inference
  of an ANN isn't (the sequentiality of its operations can't be equated
  to lived temporality).}. We hope our theoretical groundwork will
enable such transdisciplinary approaches, like there have been many for
neurophenomenology.

Applying CP to DL we have arrived at a conceptual framework that lies in
stark opposition to cognitivism and neuro-representationalism. We take
as a starting point the lived world rather than the world interpreted in
physicalist terms. From this perspective, we reject the existence of
unconscious symbols underlying our cognitive processes and propose a
theory of conscious re-presentations as re-activations of sedimented
habits that are fused to their associated processes. Finally, this
perspective also does not allow us to draw a boundary between mind and
world, like a naturalist can separate an organism from its environment.
Consciousness remains the background of our investigation, and even in
the investigated cognitive processes a ``subject of the cognition'' is
not easily found -- Merleau-Ponty for example, says that perception is
better described by ``\emph{one} perceives in me'' than ``I perceive''
(Merleau-Ponty, 1945, p. 223). Still a functional unit, the \emph{corps
propre}, can be isolated to better explain the way I adjust my grip on
the world. This has an interesting consequence: from CP the quest for an
autonomous agent, or even an AGI, is not necessarily the ultimate goal,
as we rather isolate and study particular cognitive processes (of the
\emph{corps propre}). Finally, to summarize:

\begin{enumerate}

\item
\ul{What is the world?}

The lived world, that can be investigated from first-person perspective.

\item
  \ul{What is a representation?}

A conscious representation is an act that is inextricably linked to its
corresponding process. This act is realized by \emph{re-activation} of
\emph{habits} obtained by \emph{sedimentation} of past experiences.

\item
  \ul{What is the mind?}
  
The mind is not delimitable. However, cognitive processes can be
isolated and explored through consciousness. All the cognitive processes
belong to the \emph{corps propre} and have a common objective: obtaining
an \emph{optimal grip} on the world.
\end{enumerate}

Unsurprisingly, this framework that we have obtained by applying CP to
DL, is better modeled using the mechanisms of ANNs than those of a
Turing Machine. We introduced the CP toolkit and its novel vocabulary.
We propose to consider the training of ANNs as a way to constitute an
\emph{optimal grip} on some data, just as the \emph{corps propre}
adjusts its grip on the \emph{lived world}. As such it allows the
\emph{sedimentation} of experiences into \emph{habits}. These
\emph{habits} can further be reemployed in other tasks by
\emph{re-activation}, allowing \emph{conscious re-presentations}. When
thought of as modeling processes of the \emph{corps propre}, the ANN's
inference can be seen as an \emph{intentional act} --~such as the
\emph{perceptual synthesis}, the \emph{imaging consciousness}, or even
the use of a \emph{lexical} \emph{concept}. The fact that DL is
non-decomposable is not seen as a weakness in this framework, as
conscious representations, or concepts, are inextricably linked to their
corresponding cognitive processes. In the end, the DL engineer, or
computational neuroscientist, might rely on the CP framework or the NR
framework depending on the cognitive process he is considering. If she
considers learning as happening in lived experience -- as when someone
is learning to drive a car or play chess for example -- she might rely
on the former; if she considers it as happening in the physical world --
as with the perception of a certain wavelength for example -- she could
rely on the latter.

We have proposed a new methodology to move the field of DL and
phenomenology forward as well as an alternative conceptual toolkit to
consider DL. It considers the computational from lived experience and
therefore puts on hold the question of the material basis of
consciousness. Therefore, it distinguishes itself from cognitivism and
neuro-representationalism; but also from connectionism and some forms of
enactivism (like Varela's) that think of consciousness in terms of
emergence. Notably, it has led us to a novel type mathematization that
cannot be conceived from the scientific-realism-grounded NR position.
Indeed, by considering ANNs to model the processes structuring our
experience, we oppose Galilean mathematization, criticized by Husserl,
that necessarily implies a commitment towards the existence of external
objects: for example, of physical bodies to which the law of gravitation
can apply. As such, CP opens up a novel form of computational modeling
-- that ``precedes'' the laws that govern objects -- of the processes by
which I learn to perceive, imagine, or think things. Become
mathematizable not only the physical laws applying to objects but also
the mechanisms that allow me to perceive an (intentional) object as a
delimited object distinguished from its background -- perception of
objects from which I can start to formulate the hypothesis of the
existence of an external physical world.

In such cases, and more generally for all processes considered by
computational phenomenology, what do the mathematical models tell us?
What do they correspond to? Two interpretations are possible. Either we
remain faithful to classical phenomenology by keeping the
neurophysiological source bracketed. In that case, we consider that we
are modeling mechanisms of the spontaneous mind that are neither
identical nor reducible to the physical states with which they are
correlated, but rather allow the creation of a world around us.
Furthermore, qualia are part of the starting point of our investigation
and their emergence does not need to be explained: the hard problem of
consciousness is thus avoided. Otherwise, we return to naturalism and
consider that the cognitive processes involved in the unveiling of the
physical world are themselves reducible to physical events\footnote{After
  all, the order in which we learn to conceive things does not
  necessarily entail an order of existence. As Sellars puts it: ``we
  must distinguish between primacy in the order of being and primacy in
  the order of conceiving'' (Sellars, 1971, p. 408).}. The invariant
mechanisms we have isolated by applying computational phenomenology --
that rely on reactivating sedimented lived experiences -- could be
confronted with the neural operations of the brain. Such an informed
return to the neurophysiological source could provide an opportunity to
surpass cognitivism and neuro-representationalism in modern
neuroscience\footnote{Following this second interpretation, a
  CP-informed neuroscience would reject the notion of neural
  representations, following neuroscientists that find it misleading
  (Freeman and Skarda, 1990; Brette, 2019), having identified it as the
  well-known map-territory fallacy (confusing models of reality with
  reality itself) (Korzybski, 1933). It might find common grounds with
  anti-representationalist theories such as the dynamical systems
  approach (Van Gelder, 1995; Freeman, 2000; Favela, 2021; Hipólito,
  2022) or instrumentalist accounts of the Free Energy Principle
  (Friston 2021; Andews, 2021; van Es, 2021; Parr Pezzulo and Friston,
  2022).}.

\pagebreak

\section*{Bibliography}
\begin{hangingpar}{2em}

Andrews, M. (2021). The math is not the territory: navigating the free
energy principle. Biology \& Philosophy, 36(3), 1-19.

Ashby, F. G. (2014). Multidimensional models of perception and
cognition. Psychology Press.

Baevski, A., Zhou, Y., Mohamed, A., \& Auli, M. (2020). wav2vec 2.0: A
framework for self-supervised learning of speech representations.
\emph{Advances in Neural Information Processing Systems}, \emph{33},
12449-12460.

Barbaras, R. (2020). Chapitre 19. Le vivant comme fondement originaire
de l'intentionnalité perceptive. In \emph{Naturaliser la
phénoménologie}. \url{https://doi.org/10.4000/books.editionscnrs.32336}

Bitbol, M. (2006). Une science de la conscience équitable. L'actualité
de la phénoménologie de Francisco Varela. \emph{Intellectica},
\emph{43}(1), 135-157.

Boge, F. J. (2022). Two Dimensions of Opacity and the Deep Learning
Predicament. \emph{Minds and Machines}, \emph{32}(1).
https://doi.org/10.1007/s11023-021-09569-4

Botvinick, M., Wang, J. X., Dabney, W., Miller, K. J., \& Kurth-Nelson,
Z. (2020). Deep Reinforcement Learning and Its Neuroscientific
Implications. In \emph{Neuron} (Vol. 107, Issue 4).
https://doi.org/10.1016/j.neuron.2020.06.014

Brown, T. B., Mann, B., Ryder, N., Subbiah, M., Kaplan, J., Dhariwal,
P., Neelakantan, A., Shyam, P., Sastry, G., Askell, A., Agarwal, S.,
Herbert-Voss, A., Krueger, G., Henighan, T., Child, R., Ramesh, A.,
Ziegler, D. M., Wu, J., Winter, C., \ldots{} Amodei, D. (2020). Language
models are few-shot learners. \emph{Advances in Neural Information
Processing Systems}, \emph{2020-December}.

Buckner, C. (2019). Deep learning: A philosophical introduction.
\emph{Philosophy Compass}, \emph{14}(10).
https://doi.org/10.1111/phc3.12625

Brette, R. (2019). Is coding a relevant metaphor for the brain?.
Behavioral and Brain Sciences, 42.

Chalmers, David. J. (1995). Facing Up To The Hard Problem of
Consciousness. \emph{Journal of Consciousness Studies}, \emph{2}(3).

Chemero, A. (2011). \emph{Radical embodied cognitive science}. MIT
press.

Chen, T., Kornblith, S., Norouzi, M., \& Hinton, G. (2020). A simple
framework for contrastive learning of visual representations. \emph{37th
International Conference on Machine Learning, ICML 2020},
\emph{PartF168147-3}.

Churchland, P. S., \& Sejnowski, T. J. (1990). Neural representation and
neural computation. Philosophical Perspectives, 4, 343-382.

Cohen, Y., Engel, T. A., Langdon, C., Lindsay, G. W., Ott, T., Peters,
M. A., ... \& Ramaswamy, S. (2022). Recent Advances at the Interface of
Neuroscience and Artificial Neural Networks. \emph{Journal of
Neuroscience}, \emph{42}(45), 8514-8523.

Davies, A., Veličković, P., Buesing, L., Blackwell, S., Zheng, D.,
Tomašev, N., ... \& Kohli, P. (2021). Advancing mathematics by guiding
human intuition with AI. \emph{Nature}, \emph{600}(7887), 70-74.

Devlin, J., Chang, M. W., Lee, K., \& Toutanova, K. (2019). BERT:
Pre-training of deep bidirectional transformers for language
understanding. \emph{NAACL HLT 2019 - 2019 Conference of the North
American Chapter of the Association for Computational Linguistics: Human
Language Technologies - Proceedings of the Conference}, \emph{1}.

DeVries, P. M., Viégas, F., Wattenberg, M., \& Meade, B. J. (2018). Deep
learning of aftershock patterns following large earthquakes.
\emph{Nature}, \emph{560}(7720), 632-634.

Di Paolo, E., Buhrmann, T., \& Barandiaran, X. (2017).
\emph{Sensorimotor life: An enactive proposal}. Oxford University Press.

Doerig, A., Sommers, R., Seeliger, K., Richards, B., Ismael, J.,
Lindsay, G., ... \& Kietzmann, T. C. (2022). The neuroconnectionist
research programme. \emph{arXiv preprint arXiv:2209.03718}.

Dreyfus, H. L. (1992). 2. Heidegger\textquotesingle s Hermeneutic
Realism. In \emph{The Interpretive Turn: Philosophy, Science, Culture}
(pp. 25-41). Ithaca, NY: Cornell University Press.

Dreyfus, H. L. (2002). Intelligence without representation -
Merleau-Ponty's critique of mental representation. \emph{Phenomenology
and the Cognitive Sciences}, \emph{1}(4).

Dreyfus, H. L. (2007). Why Heideggerian AI failed and how fixing it
would require making it more Heideggerian. \emph{Artificial
Intelligence}, \emph{171}(18).
\url{https://doi.org/10.1016/j.artint.2007.10.012}

Driess, D., Ha, J. S., Toussaint, M., \& Tedrake, R. (2022, January).
Learning models as functionals of signed-distance fields for
manipulation planning. In Conference on Robot Learning (pp. 245-255).
PMLR.

Eppe, M., Gumbsch, C., Kerzel, M., Nguyen, P. D., Butz, M. V., \&
Wermter, S. (2022). Intelligent problem-solving as integrated
hierarchical reinforcement learning. Nature Machine Intelligence, 4(1),
11-20.

Favela, L. H. (2021). The dynamical renaissance in neuroscience.
Synthese, 199(1), 2103-2127.

Fazi, M. B. (2021). Beyond human: Deep learning, explainability and
representation. Theory, Culture \& Society, 38(7-8), 55-77.

Fei, N., Lu, Z., Gao, Y., Yang, G., Huo, Y., Wen, J., Lu, H., Song, R.,
Gao, X., Xiang, T., Sun, H., \& Wen, J.-R. (2022). Towards artificial
general intelligence via a multimodal foundation model. \emph{Nature
Communications}, \emph{13}(1), 3094.
https://doi.org/10.1038/s41467-022-30761-2

Fodor, J. A. (1983). \emph{The modularity of mind}. MIT press.

Freeman, W. J., \& Skarda, C. A. (1990). Representations: Who needs
them?.

Freeman, W. J. (2000). \emph{How brains make up their minds}. Columbia
University Press.

Frith, C. (2007). \emph{Making up the mind: How the brain creates our
mental worlds}. Oxford: Blackwell.

Gallagher, S., \& Zahavi, D. (2020). The phenomenological mind.
Routledge.

Gallagher, S. (2017). \emph{Enactivist interventions: Rethinking the
mind}. Oxford University Press.

Gidaris, S., Singh, P., \& Komodakis, N. (2018). Unsupervised
representation learning by predicting image rotations. \emph{6th
International Conference on Learning Representations, ICLR 2018 -
Conference Track Proceedings}.

Goh, G., Cammarata, N., Voss, C., Carter, S., Petrov, M., Schubert, L.,
Radford, A., \& Olah, C. (2021). Multimodal Neurons in Artificial Neural
Networks. \emph{Distill}, \emph{6}(3).
\url{https://doi.org/10.23915/distill.00030}

Goyal, A., \& Bengio, Y. (2022). Inductive biases for deep learning of
higher-level cognition. Proceedings of the Royal Society A, 478(2266),
20210068.

Ha, D., \& Schmidhuber, J. (2018). \emph{World Models}.
\url{https://doi.org/10.5281/zenodo.1207631}

Hipólito, I. (2022). Cognition Without Neural Representation: Dynamics
of a Complex System. Frontiers in Psychology, 5472.

Hutto, D. D., \& Myin, E. (2012). \emph{Radicalizing enactivism: Basic
minds without content}. MIT press.

Husserl, E. ({[}1900{]} 2001). \emph{Logical Investigations Volume 1}.
Routledge.

Husserl, E. ({[}1931{]} 2012). \emph{Ideas: General introduction to pure
phenomenology}. Routledge.

Husserl, E. ({[}1936{]} 1970). \emph{The Crisis of European Sciences and
Transcendental Phenomenology an Introduction to Phenomenological
Philosophy}. Northwestern University Press.

Hsu, W. N., Bolte, B., Tsai, Y. H. H., Lakhotia, K., Salakhutdinov, R.,
\& Mohamed, A. (2021). Hubert: Self-supervised speech representation
learning by masked prediction of hidden units. \emph{IEEE/ACM
Transactions on Audio, Speech, and Language Processing}, \emph{29},
3451-3460.

Jumper, J., Evans, R., Pritzel, A., Green, T., Figurnov, M.,
Ronneberger, O., Tunyasuvunakool, K., Bates, R., Žídek, A., Potapenko,
A., Bridgland, A., Meyer, C., Kohl, S. A. A., Ballard, A. J., Cowie, A.,
Romera-Paredes, B., Nikolov, S., Jain, R., Adler, J., Back, T., \ldots{}
Hassabis, D. (2021). Highly accurate protein structure prediction with
AlphaFold. \emph{Nature}, \emph{596}(7873), 583--589.

Kirchhoff, M., Parr, T., Palacios, E., Friston, K., \& Kiverstein, J.
(2018). The markov blankets of life: Autonomy, active inference and the
free energy principle. \emph{Journal of the Royal Society Interface},
\emph{15}(138). \url{https://doi.org/10.1098/rsif.2017.0792}

Korzybski, A. (1933). Science and Sanity: An introduction to
non-aristotelian systems and general semantics Lakeville. Conn.:
International Non-aristotelian Library Publishing Co.

Kumar, S., Sumers, T. R., Yamakoshi, T., Goldstein, A., Hasson, U.,
Norman, K. A., Griffiths, T. L., Hawkins, R. D., \& Nastase, S. A.
(2022). Reconstructing the cascade of language processing in the brain
using the internal computations of a transformer-based language model.
\emph{BioRxiv}.

LeCun, Y., Bengio, Y., \& Hinton, G. (2015). Deep learning.
\emph{Nature}, \emph{521}(7553), 436--444.

Lees, R. B., \& Chomsky, N. (1957). Syntactic Structures.
\emph{Language}, \emph{33}(3). https://doi.org/10.2307/411160

Li, G., Duan, N., Fang, Y., Gong, M., \& Jiang, D. (2020). Unicoder-VL:
A universal encoder for vision and language by cross-modal pre-training.
\emph{AAAI 2020 - 34th AAAI Conference on Artificial Intelligence}.

Lutz, A., \& Thompson, E. (2003). Neurophenomenology: Integrating
Subjective Experience and Brain Dynamics in the Neuroscience of
Consciousness. \emph{Journal of Consciousness Studies},
\emph{10}(9--10).

MacKay, D., Shannon, C., \& McCarthy, J. (1956). \emph{Automata
studies}.

Matsuo, Y., LeCun, Y., Sahani, M., Precup, D., Silver, D., Sugiyama, M.,
Uchibe, E., \& Morimoto, J. (2022). Deep learning, reinforcement
learning, and world models. \emph{Neural Networks}, \emph{152},
267--275.

Mazzaglia, P., Verbelen, T., Çatal, O., \& Dhoedt, B. (2022). The Free
Energy Principle for Perception and Action: A Deep Learning Perspective.
Entropy, 24(2), 301.

McClelland, J. L. (2022). Capturing advanced human cognitive abilities
with deep neural networks. Trends in Cognitive Sciences.

McCulloch, W. S., \& Pitts, W. (1943). A logical calculus of the ideas
immanent in nervous activity. \emph{The Bulletin of Mathematical
Biophysics}, \emph{5}(4). https://doi.org/10.1007/BF02478259

Merleau-Ponty, M., \& Landes, D. A. ({[}1945{]} 2012).
\emph{Phenomenology of perception}. Routledge.

Metzinger, T. (2009). The ego tunnel. New York: Basic Books.

Milkowski, M. (2013). Explaining the computational mind. Mit Press.

Millet, J., Caucheteux, C., Orhan, P., Boubenec, Y., Gramfort, A.,
Dunbar, E., Pallier, C., \& King, J.-R. (2022). Toward a realistic model
of speech processing in the brain with self-supervised learning. In
\emph{arxiv.org}. \url{https://arxiv.org/abs/2206.01685}

Minsky, M. (1961). Steps Toward Artificial Intelligence. In
\emph{Proceedings of the IRE} (Vol. 49, Issue 1).
https://doi.org/10.1109/JRPROC.1961.287775

Mnih, V., Kavukcuoglu, K., Silver, D., Rusu, A. A., Veness, J.,
Bellemare, M. G., ... \& Hassabis, D. (2015). Human-level control
through deep reinforcement learning. \emph{nature}, \emph{518}(7540),
529-533.

Mordvintsev, A., Olah, C., \& Tyka, M. (2015). Inceptionism: Going
Deeper into Neural Networks. In \emph{Research Blog}.

Mrowca, D., Zhuang, C., Wang, E., Haber, N., Fei-Fei, L. F., Tenenbaum,
J., \& Yamins, D. L. (2018). Flexible neural representation for physics
prediction. Advances in neural information processing systems, 31.

Olah, C. (2015). Understanding LSTM Networks. \emph{GITHUB Colah Blog}.
\url{http://colah.github.io/posts/2015-08-Understanding-LSTMs/}

Panaccio, C. (2011). \emph{Qu'Est-Ce Qu'Un Concept?}

Panaccio, C. (2017). Ockham on concepts. In \emph{Ockham on Concepts}.
\url{https://doi.org/10.4324/9781315247809}

Parr, T., Pezzulo, G., \& Friston, K. J. (2022). Active inference: the
free energy principle in mind, brain, and behavior. MIT Press.

Perconti, P., \& Plebe, A. (2020). Deep learning and cognitive science.
Cognition, 203, 104365.

Petitot, J., \& Smith, B. (1996). Physics and the phenomenal world. In
\emph{Formal ontology} (pp. 233-253). Springer, Dordrecht.

Petitot, J. (1999). \emph{Naturalizing phenomenology: Issues in
contemporary phenomenology and cognitive science}. Stanford University
Press.

Piantadosi, S. T. (2021). The computational origin of representation.
Minds and machines, 31(1), 1-58.

Poldrack, R. A. (2021). The physics of representation. Synthese, 199(1),
1307-1325.

Putnam, H. (1967). The nature of mental states. \emph{Art, mind, and
religion}, 37-48.

Radford, A., Wook, J., Chris, K., Aditya, H., Gabriel, R., Sandhini, G.,
Sastry, G., Askell, A., Mishkin, P., Clark, J., Krueger, G., \&
Sutskever, I. (2021). Learning Transferable Visual Models From Natural
Language Supervision. \emph{OpenAI}, 47. https://github.com/openai/CLIP

Ramesh, A., Pavlov, M., Goh, G., Gray, S., Voss, C., Radford, A., Chen,
M., \& Sutskever, I. (2021). Zero-Shot Text-to-Image Generation. In
\emph{proceedings.mlr.press}. https://github.com/openai/DALL-E

Ramstead, M. J. D. (2015). Naturalizing what? Varieties of naturalism
and transcendental phenomenology. \emph{Phenomenology and the Cognitive
Sciences}, \emph{14}(4). \url{https://doi.org/10.1007/s11097-014-9385-8}

Ramstead, M. J. D., Seth, A. K., Hesp, C., Sandved-Smith, L., Mago, J.,
Lifshitz, M., Pagnoni, G., Smith, R., Dumas, G., Lutz, A., Friston, K.,
\& Constant, A. (2022). From Generative Models to Generative Passages: A
Computational Approach to (Neuro) Phenomenology. \emph{Review of
Philosophy and Psychology}.
\url{https://doi.org/10.1007/s13164-021-00604-y}

Räukur, T., Ho, A., Casper, S., \& Hadfield-Menell, D. (2022). Toward
Transparent AI: A Survey on Interpreting the Inner Structures of Deep
Neural Networks. \emph{arXiv preprint arXiv:2207.13243}.

Rosenblatt, F. (1958). The perceptron: A probabilistic model for
information storage and organization in the brain. \emph{Psychological
Review}, \emph{65}(6). https://doi.org/10.1037/h0042519

Rumelhart, D. E., Hinton, G. E., \& Williams, R. J. (1986). Learning
Internal Representations By Error Propagation. In \emph{Cognitive
Science} (Vol. 1, Issue V).

Saddler, M. R., Gonzalez, R., \& McDermott, J. H. (2021). Deep neural
network models reveal interplay of peripheral coding and stimulus
statistics in pitch perception. Nature communications, 12(1), 1-25.

Sandved-Smith, L., Hesp, C., Mattout, J., Friston, K., Lutz, A., \&
Ramstead, M. J. D. (2021). Towards a computational phenomenology of
mental action: Modelling meta-awareness and attentional control with
deep parametric active inference. \emph{Neuroscience of Consciousness},
\emph{2021}(2). \url{https://doi.org/10.1093/nc/niab018}

Sartre, J.-P., Elkaim-Sartre, Arlette., \& Webber, J. (Jonathan M.
(2004). \emph{The imaginary\,: a phenomenological psychology of the
imagination}. Routledge.

Schulman, J., Zoph, B., Kim, C., Hilton, J., Menick, J., Weng, J., ...
\& Ryder, N. (2022). ChatGPT: Optimizing language models for dialogue.

Sellars, W. (1971). Science, Sense Impressions, and Sensa: A Reply to
Cornman. \emph{The Review of Metaphysics}, \emph{24}(3), 391--447.

Silver, D. (2015). Lecture 1: Introduction to reinforcement learning.
\emph{Google DeepMind}, \emph{1}, 1-10.

Silver, D., Huang, A., Maddison, C. J., Guez, A., Sifre, L., Van Den
Driessche, G., ... \& Hassabis, D. (2016). Mastering the game of Go with
deep neural networks and tree search. \emph{nature}, \emph{529}(7587),
484-489.

Sitzmann, V., Martel, J., Bergman, A., Lindell, D., \& Wetzstein, G.
(2020). Implicit neural representations with periodic activation
functions. Advances in Neural Information Processing Systems, 33,
7462-7473.

Sloman, A. (2019). The computer revolution in philosophy: Philosophy,
science and models of mind.

Taigman, Y., Yang, M., Ranzato, M., \& Wolf, L. (2014). DeepFace:
Closing the gap to human-level performance in face verification.
\emph{Proceedings of sthe IEEE Computer Society Conference on Computer
Vision and Pattern Recognition}. https://doi.org/10.1109/CVPR.2014.220

Van Es, T. (2021). Living models or life modelled? On the use of models
in the free energy principle. Adaptive Behavior, 29(3), 315-329.

Van Gelder, T. (1995). What might cognition be, if not computation?. The
Journal of Philosophy, 92(7), 345-381.

Varela, F. J. (1996). Neurophenomenology: A methodological remedy for
the hard problem. \emph{Journal of Consciousness Studies}, \emph{3}(4).

Varela, F. J., Thompson, Evan., \& Rosch, Eleanor. (1991). \emph{The
embodied mind\,: cognitive science and human experience}. MIT Press.

Von der Malsburg, C. (1995). Binding in models of perception and brain
function. Current opinion in neurobiology, 5(4), 520-526.

Wang, X., Wang, S., Liang, X., Zhao, D., Huang, J., Xu, X., ... \& Miao,
Q. (2022). Deep reinforcement learning: a survey. IEEE Transactions on
Neural Networks and Learning Systems.

Xu, J., de Mello, S., Liu, S., Byeon, W., Breuel, T., Kautz, J., Wang,
X., \& San Diego, U. (2022). GroupViT: Semantic Segmentation Emerges
from Text Supervision. In \emph{openaccess.thecvf.com}.
https://github.com/NVlabs/GroupViT.

Yamins, D. L. K., \& DiCarlo, J. J. (2016). Using goal-driven deep
learning models to understand sensory cortex. In \emph{Nature
Neuroscience} (Vol. 19, Issue 3).
\url{https://doi.org/10.1038/nn.4244}

Yoshimi, J. (2011). Phenomenology and connectionism. \emph{Frontiers in
psychology}, \emph{2}, 288.

Zahavi, D. (2008). Phenomenology. In \emph{The Routledge companion to
twentieth century philosophy} (pp. 661-692). Routledge.

Zahavi, D. (2018). Brain, Mind, World: Predictive Coding,
Neo-Kantianism, and Transcendental Idealism. \emph{Husserl Studies},
\emph{34}(1). \url{https://doi.org/10.1007/s10743-017-9218-z}

Zhang, Y., Tino, P., Leonardis, A., \& Tang, K. (2021). A Survey on
Neural Network Interpretability. In \emph{IEEE Transactions on Emerging
Topics in Computational Intelligence} (Vol. 5, Issue 5).
\url{https://doi.org/10.1109/TETCI.2021.3100641}

\end{hangingpar}
\end{document}